\documentclass{article}
\usepackage[utf8]{inputenc}
\usepackage[margin=20mm]{geometry}
\usepackage{amsmath,amssymb}
\usepackage{algorithm}
\usepackage{algpseudocode}
\usepackage{graphicx}
\usepackage{bm}
\usepackage{color}
\usepackage{comment}
\usepackage{subcaption}
\usepackage{authblk}
\usepackage{hyperref}
\usepackage{ulem}
\usepackage{url}

\newcommand{\wt}{\widetilde}

\newcommand{\com}[1]{{\sf\color[rgb]{0,0,1}{#1}}}

\title{Bouncy particle sampler with infinite exchanging parallel tempering}
\author[1]{Yohei Saito}
\author[1]{Shun Kimura}
\author[2]{Koujin Takeda \thanks{\href{mailto:koujin.takeda.kt@vc.ibaraki.ac.jp}{koujin.takeda.kt@vc.ibaraki.ac.jp}}}

\affil[1]{Center for Mathematics and Data Science, Gunma University}
\affil[2]{Graduate School of Science and Engineering, Ibaraki University}
\date{}

\begin{document}
\maketitle
\begin{abstract}
Markov chain Monte Carlo is one of the most useful tools for obtaining samples from complex distributions, such as posterior distributions in Bayesian inference. 
Bouncy particle sampler (BPS) is a method that performs sampling using uniform linear motion and elastic collisions. 
Since BPS does not require time discretization, it is expected to be more advantageous than the widely used Hamiltonian Monte Carlo (HMC) and Langevin Monte Carlo (LMC) for distributions with large curvature changes. 
To sample more efficiently from multimodal distributions, we propose BPS with parallel tempering (BPS-PT) in the infinite exchange limit, based on LMC-PT with the infinite exchange rate proposed in a previous study. 
Using a distribution with large curvature changes, BPS-PT showed estimation accuracy comparable to HMC-PT and LMC-PT, and achieved higher ESS per computational time. 
Therefore, BPS-PT is considered to be effective for distributions with large curvature changes.
\end{abstract}

\section{Introduction}
Markov chain Monte Carlo (MCMC)~\cite{andrieu2008tutorial} is one of the most useful methods to obtain samples from complex probability distributions such as posterior distributions in Bayesian inference~\cite{afify2022power, elbatal2025classical, alsadat2024inverse, sapkota2023arc, afify2021new}. 
For continuous probability distributions, Langevin Monte Carlo (LMC)~\cite{rossky1978brownian} and Hamiltonian Monte Carlo (HMC)~\cite{neal2011mcmc} are widely used (LMC is also called Metropolis augmented Langevin algorithm (MALA)). 
For discrete probability distributions, discrete variables are updated by a transition matrix constructed to satisfy the detailed balance condition. 
Using both methods, we can obtain samples from mixed continuous-discrete distributions.


Recently, non-reversible sampling methods using a piecewise-deterministic Markov process have been proposed and are studied on theoretical aspects~\cite{peters2012rejection, bouchard2018bouncy, bierkens2020boomerang, corbella2022automatic, deligiannidis2021randomized, chin2024mcmc}, improvements~\cite{sherlock2022discrete, fearnhead2024stochastic}, and applications~\cite{sutton2022continuously, bertazzi2024piecewise}. 
Bouncy particle sampler (BPS) is a method for sampling from continuous distributions using uniform linear motion and stochastic reflection generated by an inhomogeneous Poisson process~\cite{peters2012rejection, bouchard2018bouncy}. 
By inserting transition events for discrete variables into BPS dynamics, BPS can be applied to mixed continuous-discrete distributions~\cite{bouchard2017arxiv}. 
BPS was reported to have an advantage in that it is easier to adjust simulation parameters than HMC~\cite{bouchard2018bouncy}. 
In LMC and HMC, the simulation time is discretized by $\Delta t$ while preserving the detailed balance condition. 
When the curvature (Hessian matrix) of the target distribution changes significantly, 
it is difficult to efficiently sample using a constant $\Delta t$. 
Therefore, we should adaptively change $\Delta t$ as is effectively done by Riemannian manifold HMC~\cite{girolami2011riemann} or transform the variables to obtain distributions with small curvature change (preconditioning)~\cite{seyboldt2026preconditioning}. 
The former usually increases computational costs significantly, while the latter may require complex nonlinear transformations. 
Since BPS does not require to discretize the simulation time, it may be an effective method for sampling from such distributions if we can  simulate inhomogeneous Poisson processes. 
Furthermore, partially deterministic Markov process (PDMP) samplers including BPS can be applied to distributions with the delta function~\cite{bierkens2023sticky} and heavy-tailed distributions~\cite{bierkens2019zig}. 


In some multimodal target distributions, a particle rarely moves from one mode to another, resulting in slow convergence to target distributions. 
Parallel tempering (PT) is a way to mitigate this problem by introducing temperature to facilitate inter-modal transitions~\cite{swendsen1986replica, neal1996sampling}, and is used in combination with various sampling algorithms, e.g., LMC and HMC. 
There are studies on optimizing temperature~\cite{vousden2016dynamic, zhao2024policy} and applying irreversible processes to temperature exchange~\cite{syed2022non, deng2023non}. 
Recently, it has been shown that in LMC, the higher the exchange rate of PT, the faster the convergence to the target distribution~\cite{dupuis2012infinite}. 
Furthermore, an algorithm for LMC that incorporates PT with an infinite exchange limit has been proposed~\cite{dupuis2012infinite}.

Since slow convergence in multimodal distributions is observed in the BPS, we first incorporate PT into BPS. 
Then, we propose an algorithm in the infinite exchange limit, assuming that the convergence acceleration also holds for BPS. 
Next, we present a temperature optimization that extends the method using the policy gradient in adjacent temperature exchange PT~\cite{zhao2024policy} to three or more temperature exchange PT. 
Then, we show numerical simulations using distributions where BPS is effective and where it is not. 


This paper is organized as follows. 
Section 2 reviews the BPS for continuous and mixed continuous-discrete probability distributions. 
Section 3 introduces BPS incorporating PT and proposes BPS-PT with the infinite exchange rate. 
Then, we extend temperature optimization by policy gradient to three or more temperature exchange PT. 
Section 4 performs numerical simulations and confirms that BPS-PT with infinite exchange rate is more useful to a multimodal distribution with large curvature changes than more widely used algorithms, HMC-PT and LMC-PT with a finite exchange rate. 
In Section 5, we summarize the paper. 


\section{Review of the bouncy particle sampler}
MCMC~\cite{andrieu2008tutorial} is an effective method to sample from probability distributions with unknown normalization constants. 
In particular, HMC~\cite{neal1996monte} and its improved form, No-U-Turn Sampler (NUTS)~\cite{hoffman2014no}, are widely used for continuous distributions, and MCMC with a transition matrix constructed to satisfy the detailed balance condition is employed for discrete distributions. 
BPS~\cite{peters2012rejection, bouchard2018bouncy} is a method for obtaining samples from continuous distributions by a stochastically bouncing particle moving in a straight line with constant velocity. 
In this section, following Refs.~\cite{bouchard2018bouncy, bouchard2017arxiv}, we review BPS for continuous and mixed continuous-discrete distributions. 

 \subsection{BPS for continuous distributions}

\begin{figure}[htbp]
 \centering
 \includegraphics[keepaspectratio, scale=0.5]{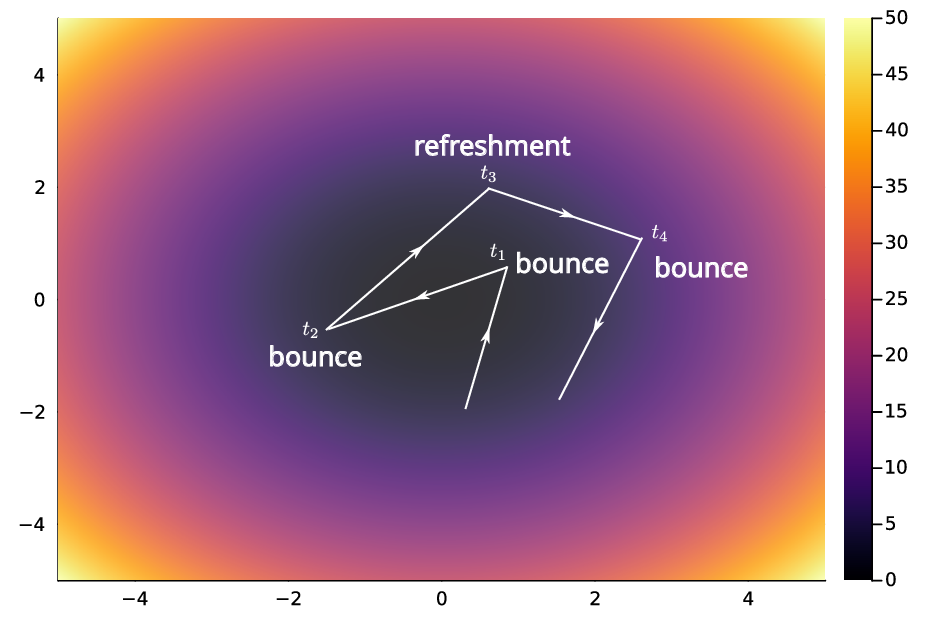}
 \caption{We show a trajectory of a particle generated by the BPS algorithm in a potential of the normal distribution. The particle bounces at $t = t_1, t_2, t_4$ and randomly changes its velocity at $t = t_3$.}
 \label{fig:potential_heatmap}
\end{figure}

Let $p(x) \propto {\rm e}^{-U(x)}$ be the target distribution, where $x \in \mathbb{R}^n$ is a continuous variable. 
Consider a particle that moves at constant velocity. 
The particle randomly changes velocity or collides elastically with the potential $U$ (Fig.~\ref{fig:potential_heatmap}). 
The elastic collisions are generated by an inhomogeneous Poisson process whose event rate is given by 
\begin{align}
\lambda^{\rm b}(x, v)
 =& \max\{0, \left< v, \nabla U(x)\right>\}, 
 \label{eq:bounce_rate}
\end{align}
where $x \in \mathbb{R}^n$ and $v \in \mathbb{R}^n$ are the position and velocity of the particle, respectively. 
After an elastic collision, the velocity of the particle changes to 
\begin{align}
 R(x) v 
 =& v - 2 \frac{\left< v, \nabla U(x)\right>}{\|\nabla U(x) \|^2} \, \nabla U(x). 
 \label{eq:bounce_velocity}  
\end{align}
BPS is a method for obtaining the target distribution $p$ as the long-term limit of the particle position distribution. 

Using a sufficient regular function $f \colon \mathbb{R}^n \times \mathbb{R}^n \to \mathbb{R}$, the infinitesimal generator of BPS is written as 
\begin{align}
 {\cal L}_{\rm BPS} f(x, v)
 =& \lambda^{\rm b}(x, v) \{ f(x, R(x) v) - f(x, v) \} 
 + \lambda^{\rm ref} \int \psi(v') \{ f(x, v') - f(x, v) \} {\rm d}v'
 + \left< v, \nabla_x f(x, v) \right>, 
 \label{eq:BPS_generator} \\
 \psi(v) =& \frac{1}{(2 \pi)^{d/2}} \exp\left[ - \frac{1}{2} \|v\|^2 \right], 
\end{align}
where $\lambda^{\rm ref} > 0$. 
The first term of Eq.~(\ref{eq:BPS_generator}) expresses a stochastic elastic collision of the particle. 
To satisfy the ergodic property, the velocity of the particle is stochastically refreshed by the standard normal distribution $\psi$ in the second term of Eq.~(\ref{eq:BPS_generator}). 
The third term of Eq.~(\ref{eq:BPS_generator}) denotes the uniform linear motion. 
From the ergodic property and 
\begin{align}
 \int {\cal L}_{\rm BPS} f(x, v) p(x) \psi(v) {\rm d}x {\rm d}v = 0,
\end{align}
$p(x) \psi(v)$ is the stationary distribution of the infinitesimal generator. 
Therefore, using the dynamics generated by ${\cal L}_{\rm BPS}$, we can obtain samples from the target distribution $p$. 
To simulate the inhomogeneous Poisson process (IPP) generated by Eq.~(\ref{eq:bounce_rate}), we employ the thinning algorithm \cite{lewis1979simulation} as follows. 
Suppose that we find an upper bound of the event rate $\wt\lambda$ in some time interval $t \in [0, T]$, 
\begin{align}
 \lambda^{\rm ref} + \lambda^{\rm b}(x + v t, v) = \lambda(t; x, v)
 \leq \wt{\lambda}(t; x, v),
 \label{eq:bounce_rate_upper_bound}
\end{align}
whose event time $\Delta t^{\rm ev}$ can be analytically evaluated as 
\begin{align}
 \ln r =& - \int_0^{\Delta t^{\rm ev}} \wt\lambda(t; x, v) {\rm d}t, \\
 r \sim& {\rm Uniform}(0, 1). 
\end{align}
For example, we choose the closest positive inflection point as the endpoint of the interval $T$ and use an upper bound, 
\begin{align}
 \wt\lambda(t; x, v) = 
 \begin{cases}
   \displaystyle \lambda(0; x, v) + \partial_t \lambda(0; x, v) \, t 
    & (\partial_t^2 \lambda(0; x, v) \leq 0) \\
   \displaystyle \lambda(0; x, v) + \frac{\lambda(T; x, v) - \lambda(0; x, v)}{T} \, t
    & (\partial_t^2 \lambda(0; x, v) > 0)
 \end{cases}.
\end{align}
Then, we determine whether the event is accepted by $\lambda(\Delta t^{\rm ev}; x, v) / \wt\lambda(\Delta t^{\rm ev}; x, v)$. 
Pseudocodes of BPS to obtain samples from the target distribution $p$ are shown in Algs.~\ref{alg:BPS}, \ref{alg:thinning}. 

BPS has an advantage that it does not require tuning the hyperparameters~\cite{bouchard2018bouncy}. 
There is another advantage in BPS. 
Unlike LMC and HMC, there is no need to discretize the simulation time into an interval $\Delta t$, which has the following advantage for distributions with large curvature change: 
In LMC (HMC), the greater the deviation from continuous-time Langevin (Hamiltonian) dynamics, the lower the acceptance rate of the Metropolis–Hastings (MH) method after the gradient update tends to be. 
To reduce the deviation from continuous-time dynamics in regions with a large curvature, $\Delta t$ should be sufficiently small. 
However, using a small $\Delta t$ in regions with small curvature reduces search efficiency. 
In this case, we adaptively change $\Delta t$ depending on the curvature or transform variables to obtain distributions with small curvature changes. 
The former method automatically adjusts $\Delta t$ by incorporating changes in curvature using motion in a Riemannian manifold~\cite{girolami2011riemann}, although it significantly increases computational costs. 
The latter method may require nonlinear transformations for complex distributions, and finding these transformations can be difficult. 
For such distributions, BPS can be a suitable choice, if we can simulate inhomogeneous Poisson processes by finding an appropriate upper bound, $\wt\lambda$. 
Furthermore, PDMP samplers including BPS are useful to sample from distributions with the delta function~\cite{bierkens2023sticky} and heavy-tailed distributions~\cite{bierkens2019zig}. 
In contrast, since updates in BPS by the thinning algorithm are often slower than gradient updates in LMC and HMC, there is little benefit to using BPS for probability distributions with small curvature changes, e.g., the normal distribution, including cases where it can be obtained through simple variable transformations. 
In Section 4, we show numerical simulations of such examples. 



\begin{algorithm}[tb]
 \caption{BPS for sampling from a continuous probability distribution}
 \label{alg:BPS}
 \begin{algorithmic}[1]
 \State Input $\Delta t^{\rm samp}, N, \lambda^{\rm ref}$
 \State Initialize $x, v$
 \State $n \leftarrow 1, \ t \leftarrow 0$
 \While{$n \leq N$}
  \State Determine event time $\Delta t^{\rm ev}$ of IPP with event rate $\lambda = \lambda^{\rm ref} + \lambda^{\rm b}$ by thinning algorithm
  \If {$t + \Delta t^{\rm ev} \geq n \Delta t^{\rm samp}$}
   \State Sample $(x + v (n \Delta t^{\rm samp} - t), v)$
   \State $n \leftarrow n + 1$
  \EndIf
  \State $t \leftarrow t + \Delta t^{\rm ev}, x \leftarrow x + v \Delta t^{\rm ev}$
  \If {$u < \lambda^{\rm ref} / \lambda(0; x, v)$ where $u \sim {\rm Uniform}(0, 1)$}
   \State $v \sim {\cal N}(0, 1)$
  \Else
   \State $v \leftarrow R(x) v$
  \EndIf
 \EndWhile
 \end{algorithmic}
\end{algorithm}

\begin{algorithm}[tb]
 \caption{Evaluation of an event time by thinning algorithm}
 \label{alg:thinning}
 \begin{algorithmic}[1]
 \State Input $x, v, \lambda$
 \State $\Delta t^{\rm ev} \leftarrow 0$
 \While{true}
  \State Find $\wt\lambda$ which satisfies $\wt\lambda(t; x, v) \geq \lambda(t; x, v), \ \forall t \in [0, T]$ for some $T$
  \State Find $\Delta t$ which satisfies $\ln r = - \int_0^{\Delta t} \wt\lambda(t; x, v) {\rm d}t$ where $r \sim {\rm Uniform}(0, 1)$
  \If {$\Delta t \leq T$}
   \If {$u \leq \lambda(\Delta t; x, v) / \wt\lambda(\Delta t; x, v)$ where $u \sim {\rm Uniform}(0, 1)$}
    \State \Return $\Delta t^{\rm ev} + \Delta t$
   \Else
    \State $x \leftarrow x + v \Delta t, \ \Delta t^{\rm ev} \leftarrow \Delta t^{\rm ev} + \Delta t$
   \EndIf
  \Else
   \State $x \leftarrow x + v T, \ \Delta t^{\rm ev} \leftarrow \Delta t^{\rm ev} + T$
  \EndIf
 \EndWhile
 \end{algorithmic}
\end{algorithm}

 \subsection{BPS for mixed continuous-discrete distributions}

Let $p(x, y) \propto {\rm e}^{-U(x, y)}$ be the target distribution, where $x \in \mathbb{R}^n$ and $y \in S$ are continuous and discrete variables, respectively ($S$ is a finite or countable set). 
To obtain samples from a continuous-discrete mixture distribution, we update the discrete variables using a transition matrix in addition to updating the continuous variables~\cite{bouchard2017arxiv, zhou2022metropolis}. 
A transition matrix $w$ is constructed so as to satisfy the balance condition, 
\begin{align} 
 \sum_{y' \neq y} w(x, y' \mid x, y) p(x, y) = \sum_{y \neq y'} w(x, y \mid x, y') p(x, y'). 
\end{align}
In many cases, the transition matrix is constructed to satisfy a stronger condition, the detailed balance condition, 
\begin{align} 
 w(x, y' \mid x, y) p(x, y) = w(x, y \mid x, y') p(x, y'). 
\end{align}

Incorporating updates of the continuous variables, the infinitesimal generator of BPS is given by 
\begin{align}
 {\cal L}'_{\rm BPS} f(x, y, v)
 =& \lambda^{\rm b}(x, y, v) \{ f(x, y, R(x, y) v) - f(x, y, v) \} 
 + \sum_{y' \neq y} \lambda^{\rm j}(x, y, y') \{ f(x, y', v) - f(x, y, v) \} \nonumber \\
 &+ \lambda^{\rm ref} \int \psi(v') \{ f(x, y, v') - f(x, y, v) \} {\rm d}v'
 + \left< v, \nabla_x f(x, y, v) \right>, 
 \label{eq:BPS_disc_generator} \\
 \lambda^{\rm j}(x, y, y')
 =& \alpha^{\rm j} w(x, y' \mid x, y), 
 \label{eq:jump_rate}
\end{align}
where $\alpha^{\rm j} > 0$ and $f \colon \mathbb{R}^n \times S \times \mathbb{R}^n \to \mathbb{R}$ is a sufficient regular function. 
Since transition events are also an IPP, we employ the thinning algorithm that uses an upper bound of the total event rate $\wt\lambda$, 
\begin{align}
 \lambda^{\rm ref} + \lambda^{\rm b}(x + v t, y, v) + \sum_{y'} \lambda^{\rm j}(x + v t, y, y')
 = \lambda(t; x, y, v) \leq \wt\lambda(t; x, y, v). 
\end{align}
If we use the MH method to construct $w$, we can always employ an upper bound $w(x, y' \mid x, y) \leq 1$. 
Thus, it is easy to find $\wt\lambda$ if we do not take into account the rejection rate of the thinning algorithm. 
We show a pseudocode in Alg.~\ref{alg:BPS_cont_disc}. 

\begin{algorithm}[tb]
 \caption{BPS for sampling from a distribution of continuous and discrete variables}
 \label{alg:BPS_cont_disc}
 \begin{algorithmic}[1]
 \State Input $\Delta t^{\rm samp}, N, \lambda^{\rm ref}, \alpha^{\rm j}$
 \State Initialize $x, y, v$
 \State $n \leftarrow 1, \ t \leftarrow 0$
 \While{$n \leq N$}
  \State Determine event time $\Delta t^{\rm ev}$ of IPP with event rate $\lambda = \lambda^{\rm ref} + \lambda^{\rm b} + \lambda^{\rm j}$ by thinning algorithm
  \If {$t + \Delta t^{\rm ev} \geq n \Delta t^{\rm samp}$}
   \State Sample $(x + v (n \Delta t^{\rm samp} - t), y, v)$
   \State $n \leftarrow n + 1$
  \EndIf
  \State $t \leftarrow t + \Delta t^{\rm ev}, x \leftarrow x + v \Delta t^{\rm ev}$
  \State $\Lambda \leftarrow (\lambda^{\rm ref}, \lambda^{\rm b}(x, y, v), (\lambda^{\rm j}(x, y, y'))_{y' \in S})$
  \State $\omega \leftarrow \Lambda / \|\Lambda\|_1$
  \State $i \sim {\rm Cat}(\omega)$
  \If {$i = {\rm ref}$}
   \State $v \sim {\cal N}(0, 1)$
  \ElsIf {$i = {\rm b}$}
   \State $v \leftarrow R(x, y) v$
  \ElsIf {$i = y'$}
   \State $y \leftarrow y'$
  \EndIf
 \EndWhile
 \end{algorithmic}
\end{algorithm}

\section{BPS-PT with infinite exchange rate}

In a multimodal target distribution, a particle may have difficulty moving from one mode to another, which results in slow convergence to the target distribution. 
PT~\cite{swendsen1986replica, neal1996sampling} has been used to solve this problem as follows. 
We introduce inverse temperatures, $1 = \beta_1 > \cdots > \beta_L > 0$, into the probability distribution as $p^{\beta_i}(x, y) \propto {\rm e}^{-\beta_i U(x, y)}$, move the particles in each distribution in parallel, and then exchange the positions of the particles. 
As the inverse temperature decreases, the probability distribution flattens, and a particle becomes less trapped in one mode. 
By exchanging the positions of the particles between different inverse temperatures, the particle at $\beta_1$ can move from one mode to another more frequently, which accelerates the convergence to the target distribution. 
In LMC, it was shown that the convergence accelerates as the exchange rate increases, and LMC with infinite exchange PT was proposed~\cite{dupuis2012infinite}. 
Assuming the same property holds for BPS, we introduce PT with finite and infinite exchange rates to BPS. 
Then, we extend a temperature optimization method by policy gradient~\cite{zhao2024policy} to PT exchanging three or more temperatures. 

%

 \subsection{BPS with PT}

In many cases, PT exchanges particles between adjacent inverse temperatures using a homogeneous Poisson process. 
Since the particle positions change discontinuously before and after an exchange, a piecewise continuous trajectory cannot be obtained in the limit of infinite exchange rate. 
Instead of fixing the inverse temperatures and exchanging the particle positions, the authors of Ref.~\cite{dupuis2012infinite} fix the particle positions and exchange the inverse temperatures. 
In this situation, we can obtain continuous trajectories in the limit of the infinite exchange rate. 
Furthermore, they use the permutation group $S_L$ to exchange inverse temperatures not only between adjacent particles but also for all particles. 

We introduce this kind of PT to BPS. 
The event rate that a particle set $\{(x_i, y_i, v_i)\}_{i=1}^L$ changes the inverse temperature distribution from $(\beta_{\sigma(1)}, \ldots, \beta_{\sigma(L)})$ to $(\beta_{\sigma'(1)}, \ldots, \beta_{\sigma'(L)})$ is given by $\alpha^{\rm s} g_{\sigma \sigma'}(x, y)$, where 
\begin{align}
 g_{\sigma \sigma'}(x, y) = \min \left\{ 1, \frac{\prod_{i=1}^L p(x_i, y_i)^{\beta_{\sigma'(i)}}}{\prod_{i=1}^L p(x_i, y_i)^{\beta_{\sigma(i)}}} \right\}. 
\end{align}
Using a sufficiently regular function $f \colon \mathbb{R}^{nL} \times S^L \times \mathbb{R}^{nL} \times S_L \to \mathbb{R}$, the infinitesimal generator of BPS-PT with a finite exchange rate is written as 
\begin{align}
 {\cal L}_{\rm BPSPT} f(x, y, v, \sigma)
 =& \sum_{i=1}^L \biggl[ 
  \lambda^{\rm b}_{\sigma(i)}(x_i, y_i, v_i) \{ f(x, y, R_i(x_i, y_i) v, \sigma) - f(x, y, v, \sigma) \} \nonumber \\ 
 &+ \sum_{y_i' \neq y_i} \lambda^{\rm j}_{\sigma(i)}(x_i, y_i, y'_i) \{ f(x, R'_i(y'_i) y, v, \sigma) - f(x, y, v, \sigma) \} \nonumber \\
 &+ \lambda^{\rm ref} \int \psi(v'_i) \{ f(x, y, R'_i(v'_i) v, \sigma) - f(x, y, v, \sigma) \} {\rm d}v'_i \biggr] \nonumber \\
 &+ \alpha^{\rm s} \sum_{\sigma' \neq \sigma} g_{\sigma, \sigma'}(x, y) \{ f(x, y, v, \sigma') - f(x, y, v, \sigma) \} 
 + \left< v, \nabla_x f(x, y, v, \sigma) \right> , 
 \label{eq:BPS-PT_generator} 
\end{align}
where an elastic collision, change of the discrete variable, and refresh of velocity of $i$-th particle is expressed by 
\begin{align}
 R_i(x_i, y_i) v
 =& (v_1, \ldots, v_{i-1}, R(x_i, y_i) v_i, v_{i+1}, \ldots, v_L), 
 \label{eq:BPS-PT_bounce} \\
 R'_i(y'_i) y
 =& (y_1, \ldots, y_{i-1}, y'_i, y_{i+1}, \ldots, y_L), \\ 
 R'_i(v'_i) v
 =& (v_1, \ldots, v_{i-1}, v'_i, v_{i+1}, \ldots, v_L). 
\end{align}
Note that from Eq.~(\ref{eq:bounce_velocity}), the velocity after an elastic collision is not affected by the inverse temperature. 
The reflection and transition rates are given by 
\begin{align}
 \lambda^{\rm b}_{\sigma(i)}(x_i, y_i, v_i)
 =& \max\{0, \left< v_i, \beta_{\sigma(i)} \nabla_x U(x_i, y_i)\right>\}
 = \beta_{\sigma(i)} \lambda^{\rm b}(x_i, y_i, v_i), 
 \label{eq:BPS-PT_bounce_rate} \\
 \lambda^{\rm j}_{\sigma(i)}(x_i, y_i, y'_i)
 =& \alpha^{\rm j} w_{\sigma(i)}(x_i, y'_i \mid x_i, y_i), 
 \label{eq:BPS-PT_jump_rate} \\
 \sum_{y'_i \neq y_i} w_{\sigma(i)}(x_i, y'_i \mid x_i, y_i) p(x_i, y_i)^{\beta_{\sigma(i)}}
 =& \sum_{y_i \neq y'_i} w_{\sigma(i)}(x_i, y_i \mid x_i, y'_i) p(x_i, y'_i)^{\beta_{\sigma(i)}}. 
 \label{eq:BPS-PT_balance_condition}
\end{align}

BPS-PT with a finite exchange rate can be simulated by the IPP whose event rate $\lambda$ is given by 
\begin{align}
 \lambda(t; x, y, v, \sigma)
 = \sum_{i=1}^L \left[\lambda^{\rm b}_{\sigma(i)}(x_i + v_i t, y_i, v_i) 
 + \sum_{y'_i \neq y_i} \lambda^{\rm j}_{\sigma(i)}(x_i + v_i t, y_i, y'_i) \right]
 + \alpha^{\rm s} \sum_{\sigma' \neq \sigma} g_{\sigma, \sigma'}(x + v t, y). 
\end{align}
Due to the symmetry of the particles, the stationary distribution is 
\begin{align}
 P_{\rm st}(x, y, v) \propto 
 \frac{1}{L!} \sum_{\sigma \in S_L} \prod_{i=1}^L p(x_i, y_i)^{\beta_{\sigma(i)}} \psi(v_i). 
 \label{eq:BPS-PT_stationary}
\end{align}
Similarly to Ref.~\cite{dupuis2012infinite}, to obtain the same sampling as with PT that exchanges the particle positions, we rearrange the particle positions using the temperature distribution at each time, $z(t)$, as 
\begin{align}
 \eta_{\alpha^{\rm s}}(T) 
  =& \frac{1}{T} \int_0^T \sum_{\sigma \in S_L} 1_{z(t) = \sigma} \delta_{(\sigma^{-1}(x(t)), \sigma^{-1}(y(t)), v(t))} {\rm d}t, 
  \label{eq:emp_dist_finite} \\
 \sigma^{-1}(x(t))
  =& (x_{\sigma^{-1}(1)}(t), \ldots, x_{\sigma^{-1}(L)}(t)), 
\end{align}
which converges to the target distribution $\prod_{i=1}^L p(x_i, y_i)^{\beta_i} \psi(v_i)$ in $T \to \infty$. 
In this way, we can obtain samples from the target distribution.

 \subsection{Infinite exchange rate limit of BPS-PT}

For LMC, using the rate function of the large deviation principle, it was shown that the higher the exchange rate $\alpha^{\rm s}$ of PT, the faster the convergence to the target distribution~\cite{dupuis2012infinite}. 
It was proved that, if several conditions are satisfied, the rate function of an empirical measure in PDMP samplers including BPS can be written in the Donsker-Varadhan variational formulation~\cite{bierkens2021large}. 
However, when the spatial dimension $n > 1$, solving the variation and obtaining the explicit rate function is an open question~\cite{bierkens2021large}. 
Therefore, in this paper, we do not show the usefulness of taking an infinite exchange rate in BPS-PT using the rate function and 
assume that the same holds for BPS. 
Referring to Ref.~\cite{dupuis2012infinite}, the infinitesimal generator of BPS-PT with infinite exchange rate is written as 
\begin{align}
 {\cal L}^\infty_{\rm BPSPT} f(x, y, v)
 =& \sum_{i=1}^L \biggl[ \lambda^{\infty, \rm b}_i(x_i, y_i, v_i) \{ f(x, y, R_i(x_i, y_i) v) - f(x, y, v) \} \nonumber \\
 &+ \sum_{y_i' \neq y_i} \lambda^{\infty, \rm j}_i(x_i, y_i, y'_i) \{ f(x, R'_i(y'_i) y, v) - f(x, y, v) \} \nonumber \\
 &+ \lambda^{\rm ref} \int \psi(v'_i) \{ f(x, y, R'_i(v'_i) v) - f(x, y, v) \} {\rm d}v'_i \biggr]
 + \left< v, \nabla_x f(x, y, v) \right>, 
 \label{eq:BPS-PT-inf_generator} 
\end{align}
where the reflection and transition rates are given by 
\begin{align}
 \lambda^{\infty, \rm b}_i(x_i, y_i, v_i)
 =& \sum_{\sigma \in S_L} \beta_{\sigma(i)} \omega_\sigma(x, y) \lambda^{\rm b}(x_i, y_i, v_i), 
 \label{eq:bounce_rate_inf} \\
 \lambda^{\infty, \rm j}_i(x_i, y_i, y'_i)
 =& \sum_{\sigma \in S_L} \omega_\sigma(x, y) w_{\sigma(i)}(x_i, y'_i \mid x_i, y_i), 
 \label{eq:jump_rate_inf} \\
 \omega_\sigma(x, y) 
 =& \frac{\prod_{i=1}^L p(x_i, y_i)^{\beta_{\sigma(i)}}}{\sum_{\sigma' \in S_L} \prod_{i=1}^L p(x_i, y_i)^{\beta_{\sigma'(i)}}}. 
 \label{eq:temperature_weight} 
\end{align}

An intuitive explanation for obtaining the event rates at the infinite exchange rate, Eqs.~(\ref{eq:bounce_rate_inf}), (\ref{eq:jump_rate_inf}), and (\ref{eq:temperature_weight}), is as follows. 
At a finite exchange rate, the probability that $i$-th particle undergoes reflection, transition, or refreshment within an infinitesimal time $\delta t$ is 
\begin{align}
 \sum_{\sigma \in S_L} \delta_{z, \sigma} \left( \beta_{z(i)} \lambda^{\rm b}(x_i, y_i, v_i) + \sum_{y' \in S} \lambda^{\rm j}_{z(i)}(x_i, y'_i \mid x_i, y_i) \right) \delta t + \lambda^{\rm ref} \delta t + o(\delta t). 
\end{align}
When the exchange rate is sufficiently large, a separation of time scales arises between the fast variable $z$ and the slow variables $x, y, v$. 
Within infinitesimal time $\delta t$, $x, y, v$ are almost fixed, and $1_{z=\sigma} \delta t$ can be approximated by $\omega_\sigma(x, y) \delta t$. 
As the exchange rate increases, $1_{z=\sigma} \delta t$ approaches $\omega_\sigma(x, y) \delta t$, giving us Eqs.~(\ref{eq:bounce_rate_inf}) and (\ref{eq:jump_rate_inf}) for the bounce and transition rates. 

The stationary distribution is the same as that of the BPS-PT with a finite exchange rate and is given by Eq.~(\ref{eq:BPS-PT_stationary}). 
Similarly to Eq.~(\ref{eq:emp_dist_finite}), the distribution of the position of the particles after rearranging, 
\begin{align}
 \eta_\infty(T) 
  =& \frac{1}{T} \int_0^T \sum_{\sigma \in S_L} \omega_\sigma(x(t), y(t)) \delta_{(\sigma^{-1}(x(t)), \sigma^{-1}(y(t)), v(t))} {\rm d}t,
\end{align}
converges to the target distribution in $T \to \infty$. 

As the number of inverse temperatures $L$ increases, the number of elements in the permutation group increases enormously, making it impossible to take the sum over the elements in the permutation group. 
Therefore, as in Ref.~\cite{dupuis2012infinite}, we use a lightweight version that uses generating subgroups of $S_L$. 
For example, we can find two generating subgroups, $G$ and $G'$, as follows. 
First, we divide $\{1, \ldots, L\}$ into two different partitions that satisfy 
\begin{align}
 & \{1, \ldots, L\} = \bigcup_{i=1}^N A_i = \bigcup_{i=1}^N B_i, \\ 
 & A_1 \subsetneq B_1, \\  
 & B_{i-1} \cap A_i \neq \emptyset \ \ (i = 2, \ldots, N), \\
 & A_i \cap B_i \neq \emptyset \ \ (i = 2, \ldots, N), \\ 
 & B_N \subsetneq A_N . 
\end{align}
Then consider the products of the permutation groups within the subsets, $G = S_{|A_1|} \times \ldots \times S_{|A_N|}$ and $G' = S_{|B_1|} \times \ldots \times S_{|B_N|}$. 
Since any element in $S_L$ can be expressed as a finite product of elements in $G$ and $G'$, they are generating subgroups of $S_L$. 
Instead of $S_L$, we alternately use $G$ and $G'$ to exchange the inverse temperatures at each interval $t_\beta$. 

The reflection and transition rate for $G$ are given as follows. 
First, we divide $\{1, \ldots, L\}$ into $A_1, \ldots, A_N$. 
We introduce an event rate matrix $\Lambda$, 
\begin{align}
 \Lambda(t; x, y, v; A) =& 
 \begin{pmatrix}
  \Lambda_1(t; x, y, v; A) \\
  \vdots \\
  \Lambda_L(t; x, y, v; A) \\
 \end{pmatrix}, \\
 \Lambda_i(t; x, y, v; A)
 =& (\lambda^{\rm ref}, \lambda^{\infty, {\rm b}}_i(x_i + v_i t, y_i, v_i; A), (\lambda^{\infty, {\rm j}}_i(x_i + v_i t, y_i, y'; A))_{y' \in S}). 
\end{align}
Suppose $i \in A_l$. 
Then, the bounce and transition rates, $\lambda^{\infty, {\rm b}}_i$ and $\lambda^{\infty, {\rm j}}_i$, are given by 
\begin{align}
 \lambda^{\infty, \rm b}_i(x_i, y_i, v_i; A)
 =& \sum_{\sigma \in S_{|A_l|}} \beta_{\sigma(i)} \omega_{l, \sigma}(x, y) 
  \lambda^{\rm b}(x_i, y_i, v_i), 
 \label{eq:bounce_rate_inf_A} \\
 \lambda^{\infty, \rm j}_i(x_i, y_i, y'_i; A)
 =& \alpha^{\rm j} \sum_{\sigma \in S_{|A_l|}} \omega_{l, \sigma}(x, y) w_{\sigma(i)}(x_i, y'_i \mid x_i, y_i), 
 \label{eq:jump_rate_inf_A} \\ 
 \omega_{l, \sigma}(x, y; A) 
 =& \frac{\prod_{j \in A_l} p(x_j, y_j)^{\beta_{\sigma(j)}}}{\sum_{\sigma' \in S_{|A_l|}} \prod_{j \in A_l} p(x_j, y_j)^{\beta_{\sigma'(j)}}}. 
 \label{eq:temperature_weight_inf}
\end{align}

If the transition matrix $w_{\sigma(i)}$ is constructed by the MH method, 
we can find an upper bound of the event rate as follows. 
Eq.~(\ref{eq:bounce_rate_inf_A}) satisfies 
\begin{align}
 \lambda^{\infty, \rm b}_i(x_i, y_i, v_i; A)
 \leq& \max_{j \in A_l} \beta_j \, \lambda^{\rm b}(x_i, y_i, v_i). 
 \label{eq:bounce_rate_inf_2}
\end{align}
From $w_{\sigma(i)} \leq 1$, Eq.~(\ref{eq:jump_rate_inf_A}) satisfies 
\begin{align}
  \lambda^{\infty, \rm j}_i(x_i, y_i, y'_i; A)
 \leq& \alpha^{\rm j} \sum_{\sigma \in S_{|A_l|}} \omega_{l, \sigma}(x, y) = \alpha^{\rm j}. 
 \label{eq:jump_rate_inf_2} 
\end{align}
Hence, using an upper bound of $\lambda^{\rm b}$ in $t \in [0, T]$, 
\begin{align}
 \lambda^{\rm b}(x_i + v_i t, y_i, v_i) \leq \wt\lambda^{\rm b}(x_i + v_i t, y_i, v_i), 
\end{align}
we obtain an upper bound of the event rate, 
\begin{align}
 & \lambda^{\rm ref} + \lambda^{\infty, \rm b}_i(x_i + v_i t, y_i, v_i; A) + \lambda^{\infty, \rm j}_i(x_i + v_i t, y_i, v_i; A)
 = \lambda_i(t; x_i, y_i, v_i, A) \\
 \leq& \lambda^{\rm ref} + \max_{j \in A_l} \beta_j \, \wt \lambda^{\rm b}(x_i + v_i t, y_i, v_i) + \alpha^{\rm j} = \wt\lambda_i(t; x_i, y_i, v_i, A). 
 \label{eq:event_rate_inf_upper_bound}
\end{align}
In addition, we construct a submatrix $\Lambda^{(A_l)}$ by extracting row vectors in $\Lambda$ that belong to $A_l$. 
Then, the total event rate of the particles in $A_l$ can be bounded by 
\begin{align}
 \|\Lambda^{(A_l)}(t; x, y, v, A)\|_1 \leq \sum_{i \in A_l} \wt\lambda_i(t; x_i, y_i, v_i, A). 
\end{align}

After time $t_\beta$ has passed, we determine the inverse temperature ordering of the particles using $\prod_{l=1}^N \omega_{l, \sigma_l}(x, y)$ and switch to the other subgroup $G'$. 
The shorter the subgroup switching time $t_\beta$, the closer the method will be to use $S_L$, although the computational cost increases. 
Thus, we should set $t_\beta$ appropriately short, depending on the problem. 
After switching the subgroups $n_s$ times, we evaluate $\prod_{l=1}^N \omega_{l, \sigma_l}(x, y)$ to determine the inverse temperature ordering and can obtain a sample from each $p^{\beta_i} \ (i=1, \ldots, L)$. 
To reduce the number of times the temperature order is calculated, we set the sampling interval to $n_s t_\beta$. 
Pseudocodes are shown in Algs.~\ref{alg:BPS_PT} and \ref{alg:BPS_PT_t_beta}.

\begin{algorithm}[tb]
 \caption{BPS-PT with infinite exchange rate}
 \label{alg:BPS_PT}
 \begin{algorithmic}[1]
 \State Input $n_{\rm s}, t_\beta, N, G, G', \lambda^{\rm ref}, \alpha^{\rm j}, L$
 \State Initialize $\{(x_i, y_i, v_i)\}_{i=1}^L$
 \State Determine $\beta_1, \ldots, \beta_L$ by policy gradient
 \State $n \leftarrow 1, \ m \leftarrow 1, \ H \leftarrow G$
 \While{$n \leq N$}
  \If{$H = G$}
   \For{$A \in \{A_1, \ldots, A_N\}$}
    \State Generate $\{(x_i, y_i, v_i)\}_{i \in A}$ after $t_\beta$
   \EndFor
   \State $H \leftarrow G'$
  \Else
   \For{$B \in \{B_1, \ldots, B_N\}$}
    \State Generate $\{(x_i, y_i, v_i)\}_{i \in B}$ after $t_\beta$
   \EndFor
   \State $H \leftarrow G$
  \EndIf
  \State $m \leftarrow m + 1$
  \If{$m = n_{\rm s}$}
   \State Sample $\{(x_i, y_i, v_i)\}_{i=1}^L$
   \State $n \leftarrow n + 1, \ m \leftarrow 1$
  \EndIf
 \EndWhile
 \end{algorithmic}
\end{algorithm}

\begin{algorithm}[tb]
 \caption{Time translation of BPS-PT in a subset $A_l$}
 \label{alg:BPS_PT_t_beta}
 \begin{algorithmic}[1]
 \State Input $t_\beta, \{\beta_i\}_{i \in A_l}, \lambda^{\rm ref}, \alpha^{\rm j}, \{(x_i, y_i, v_i)\}_{i \in A_l}$
 \State $t \leftarrow 0$
 \While{$t < t_\beta$}
  \State Determine event time $\Delta t^{\rm ev}$ of IPP with event rate $\lambda = \sum_{i \in A_l} (\lambda^{\rm ref} + \lambda^{\rm b}_i + \lambda^{\rm j}_i)$ by thinning algorithm
  \If {$t + \Delta t^{\rm ev} \geq t_\beta$}
   \State $x_i \leftarrow x_i + v_i (t_\beta - t)$ for all $i \in A_l$
   \State $\sigma \sim \omega_{l, \cdot}(x, y)$
   \State \Return $\{(x_{\sigma^{-1}(i)}, y_{\sigma^{-1}(i)}, v_i)\}_{i \in A_l}$
  \EndIf
  \State $t \leftarrow t + \Delta t^{\rm ev}, x_i \leftarrow x_i + v_i \Delta t^{\rm ev}$ for all $i \in A_l$
  \State $\Lambda_i \leftarrow (\lambda^{\rm ref}, \lambda^{\infty, {\rm b}}_i(x, y, v, A), (\lambda^{\infty, {\rm j}}_i(x, y, y', A))_{y' \in S})$ for all $i \in A_l$
  \State $\omega \leftarrow \Lambda^{(A_l)} / \|\Lambda^{(A_l)}\|_1$
  \State $(j, k) \sim {\rm Cat}(\omega)$
  \If {$k = {\rm ref}$}
   \State $v_j \leftarrow {\cal N}(0, 1)$
  \ElsIf {$k = {\rm b}$}
   \State $v_j = R(x_j, y_j) v_j$
  \ElsIf {$k = y'$}
   \State $y_j \leftarrow y'$
  \EndIf
 \EndWhile 
 \end{algorithmic}
\end{algorithm}

 \subsection{Tuning temperatures}
For PT only exchanging adjacent temperatures, there is a method to optimize the temperatures so that the exchange rate between each temperature is constant~\cite{vousden2016dynamic}. 
However, for PT exchanging three or more temperatures, there is no guideline for determining the exchange rate between non-adjacent temperatures. 
Furthermore, the relationship between efficient exploration, which is the goal of PT, and exchange frequency is not clear. 
Recently, in PT exchanging adjacent temperatures, the policy gradient was applied to optimize temperatures~\cite{zhao2024policy}. 
Therefore, we extend this method to PT exchanging three or more temperatures and apply it to optimize the temperatures of BPS-PT. 

First, we briefly explain the optimization by policy gradient in Ref.~\cite{zhao2024policy}. 
The action is inverse temperature difference $D = (d_1, \ldots, d_{L-1}), \ d_i = \ln \beta_i - \ln \beta_{i+1}$, and we do not consider state transition, that is, we deal with only one state. 
Thus, the policy function depends only on the action $\pi_\theta \colon \mathbb{R}_+^{L-1} \to \mathbb{R}$, and the parameter $\theta$ is determined by the policy gradient. 
An efficiency of particle exchange can be considered as follows: 
\begin{enumerate}
 \item[(i)] A particle-exchange effectively expands a search range, when a high-temperature particle is farther away from an area where a low-temperature particle has already explored. 
 \item[(ii)] A search range is not much expanded by an exchange of particles that are close together. 
\end{enumerate}
These properties can be expressed by the distance between particles to be exchanged as 
\begin{align}
 R_2(i, t; m, d_i) = \frac{1}{m} \sum_{j = 1}^m d(x^{(t)}_{i+1}, x^{(t-j)}_i), 
 \label{eq:SMD_adjacent}
\end{align}
where $x_{i+1}^{(t)}$ is the position of $(i + 1)$-th particle at time $t$, $d$ is the distance, and $m$ represents a past search area of $i$-th particle to be taken into account. 
Eq.~(\ref{eq:SMD_adjacent}) is called the swap mean distance (SMD) and the reward function $r_2$ is given by 
\begin{align}
 r_2(D) = \sum_{i=1}^{L-1} \sum_{t=1}^T R_2(i, t; m, d_i), 
\end{align}
where $T$ is the episode length of the reinforcement learning. 
It was experimentally shown that as the SMD increases, the autocorrelation time between samples decreases. 

We introduce particle exchanges among three or more temperatures in the SMD. 
When we divide the particles into subsets, $A_1, \ldots, A_N$ and $B_1, \ldots, B_N$, the set of particles at high temperatures to be swapped by $i$-th particle through one exchange is
\begin{align}
 C_i =& \{ j \in A_l \cup B_{l'} \mid i \in A_l \cup B_{l'}, i < j \}.
 \label{eq:i_pairs}
\end{align}
The SMD of $i$-th particle, $R_{A, B}$ and the reward function $r_{A, B}$ can be written by 
\begin{align}
 R_{A,B}(i, t; m, D) 
  =& \sum_{j \in C_i} \sum_{\tau = 1}^m d(x^{(t)}_j, x^{(t-j)}_i), \\
 r_{A, B}(D) 
  =& \frac{1}{m} \sum_{i=1}^{L-1} \sum_{t=1}^T R_{A,B}(i, t; m, D). 
 \label{eq:SMD}
\end{align}
To restrict the difference of adjacent temperatures to $(0, D_{\rm max})$, we use the beta distribution as the policy function with the parameter $\theta = \{(a_i, b_i)\}_{i=1}^{L-1}$, 
\begin{align}
 d_i =& D_{\rm max} c_i, \\ 
 c_i \sim& {\rm Beta}(a_i, b_i). 
\end{align}
As in Ref.~\cite{zhao2024policy}, we employ the REINFORCE algorithm to evaluate the policy gradient.

\section{Numerical simulation}

To discuss an advantage of BPS-PT, we employed three target distributions. 
Two of them have high curvature variations: a mixture of Gamma distributions with uniform circular distributions (MGC), and a mixture of Neal’s funnel (MNF)~\cite{neal2003slice}. 
One of them has low curvature variations: the Gaussian mixture model used in Ref.~\cite{zhou2020mixed}. 
We applied HMC-PT, LMC-PT, regular BPS, and BPS-PT with infinite exchange rate to these models and compared the results. 
Although the MGC and MNF can be easily transformed into distributions with low curvature changes, we assume that finding an appropriate transformation is difficult; therefore, we will use these distributions as they are, without performing any variable transformations.
In the following, we simply write that the term “BPS-PT” refers to the version with infinity exchange rate. 
As performance metrics for sampling methods, we employed the Wasserstein distance $W_1$ for the continuous variable, the Kullback-Leibler divergence (KLD) for the discrete variable, and the effective sample size (ESS). 
The ESS was evaluated by the minimum of the ESS of each component, 
\begin{align}
 {\rm ESS} =& \min_k {\rm ESS}_k = \min_k \frac{1}{1 + 2 \sum_{\tau = 1}^T \rho_\tau(x_{1,k})},
\end{align}
where $x_{1,k}$ denotes the $k$-th component of the continuous variable at $\beta_1 = 1$, and $\rho_\tau(x_{1,k})$ is its autocorrelation function of lag $\tau$.

For all methods, the transition matrix of the discrete variables is constructed from the MH method with a uniform proposal distribution. 
For all simulations, we used $10$ inverse temperatures and exchanged the particles or temperatures by using generating subgroups of $S_{10}$. 
Partitions of inverse temperatures were $\{\{1, \ldots, 4\}, \{5, \ldots, 8\}, \{9, 10\}\}$ and $\{\{1, 2\}, \{3, \ldots, 6\}, \{7, \ldots, 10\}\}$. 
In BPS-PT, we optimized the temperatures using the policy gradient and fixed other parameters in Alg.~\ref{alg:BPS_PT} as $\lambda^{\rm ref} = 0.1, \alpha^{\rm j} = 1, t_\beta = 0.1, n_s = 10$. 
Upper bounds of the bounce rate for the MGC and MNF are written in the Appendix. 

In HMC-PT and LMC-PT, a sampling interval and a temperature exchange interval are fixed as $\Delta t^{\rm samp} = \Delta t^{\rm exc} = 5.0$, and the number of gradient updates was given by $\lceil \Delta t^{\rm samp} / \Delta t_i \rceil$, where $\Delta t_i$ is the discrete time width at $\beta_i$. 
(To reduce computation time, we limited the maximum number of gradient updates for the MGC to $2 \times 10^3$ for HMC-PT and $10^3$ for LMC-PT, respectively.) 
We first optimized the discrete time width $\Delta t_1$, optimized the temperatures using the policy gradient, and then optimized $\Delta t_i$ for $2 \leq i \leq L$. 
Each $\Delta t_i$ is optimized by the dual averaging method~\cite{hoffman2014no} to approach the optimal acceptance rate in the MH step ($0.65$ for HMC~\cite{neal2011mcmc, beskos2013optimal} and $0.574$ for LMC~\cite{beskos2013optimal}, respectively). 

To match the computation time for optimization, the episode length used in the policy gradient of BPS-PT was set to $1/10$ of that of HMC-PT and LMC-PT ($1,000$ for HMC-PT and LMC-PT). 
All experiments were run on a machine with Intel(R) Core(TM) i7-14700 CPU, and 32 GB RAM. 

 \subsection{Mixture of Gamma distributions with uniform circular distributions}
Consider a Gamma distribution whose probability is concentrated on a circle, 
\begin{align}
 \wt p(r, \theta) = {\rm Gamma}(k, \theta \mid r) \, {\rm Uniform}(0, 2 \pi \mid \theta)
 = \frac{1}{\Gamma(k) \, \theta^k} \, r^{k-1} \, {\rm e}^{- r / \theta} \, \frac{1}{2 \pi}, 
\end{align}
and transform it to the Cartesian coordinate, 
\begin{align}
 p_{\rm GC}(x \mid k, \theta) = \frac{1}{2 \pi \, \Gamma(k) \, \theta^k} \, \|x\|^{k - 2} {\rm e}^{- \|x\| / \theta}. 
\end{align}
The probability distribution function (pdf) of the MGC is given by 
\begin{align}
 p(x, y) = a \delta_{y, 0} \, p_{\rm GC}(x \mid k_0, \theta_0) + (1 - a) \delta_{y, 1} \, p_{\rm GC}(x \mid k_1, \theta_1). 
 \label{eq:pdf_MGC}
\end{align}

We used $a = 0.3, (k_0, \theta_0) = (10^3, 10^{-3}), (k_1, \theta_1) = (4 \times 10^3, 0.5 \times 10^{-3})$. 
Since most of the probability is concentrated in the vicinity of two circumferences, $\|x\| = 1$ and $\|x\| = 2$, $\Delta t_1$ of HMC-PT and LMC-PT were small to avoid a low acceptance rate. 
Due to the trade-off between the number of gradient updates and the sampling interval, the number of updates was limited to $2 \times 10^3$ for HMC-PT and $10^3$ for LMC-PT, respectively. 
The computation times to optimize the parameters and samplings are shown in Tab.~\ref{tab:MGC_time}. 
BPS-PT has to evaluate the temperature mixing rate $\omega_{l, \sigma}$ in Eq.~(\ref{eq:temperature_weight_inf}) to obtain an event rate of the IPP. 
For the setting of this simulation, $4! = 24$ times computation was performed for each partition. 
Hence, the sampling time per sample of BPS-PT is $20$ times longer than that of BPS. 

\begin{figure}[tbp]
  \centering
  \begin{minipage}{.3\linewidth}
    \centering
    \includegraphics[width=\linewidth]{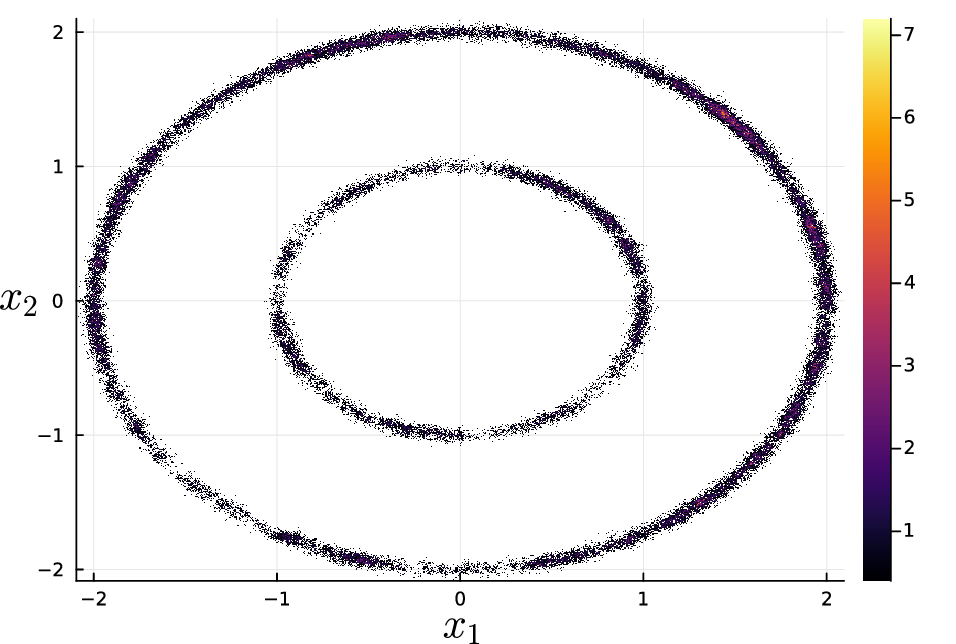}
    \subcaption{HMC-PT}
    \label{fig:MGC_HMC_histogram}
  \end{minipage}
  \hspace{5mm}
  \begin{minipage}{.3\linewidth}
    \centering
    \includegraphics[width=\linewidth]{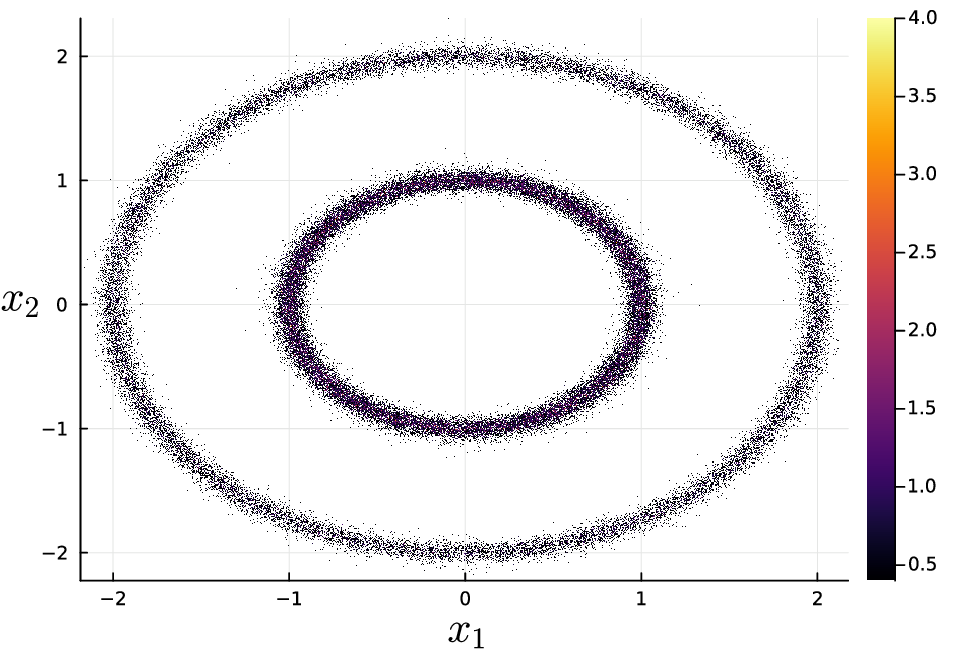}
    \subcaption{LMC-PT}
    \label{fig:MGC_LMC_histogram}
  \end{minipage}
  \vspace{3mm}  
  \begin{minipage}{.3\linewidth}
    \centering
    \includegraphics[width=\linewidth]{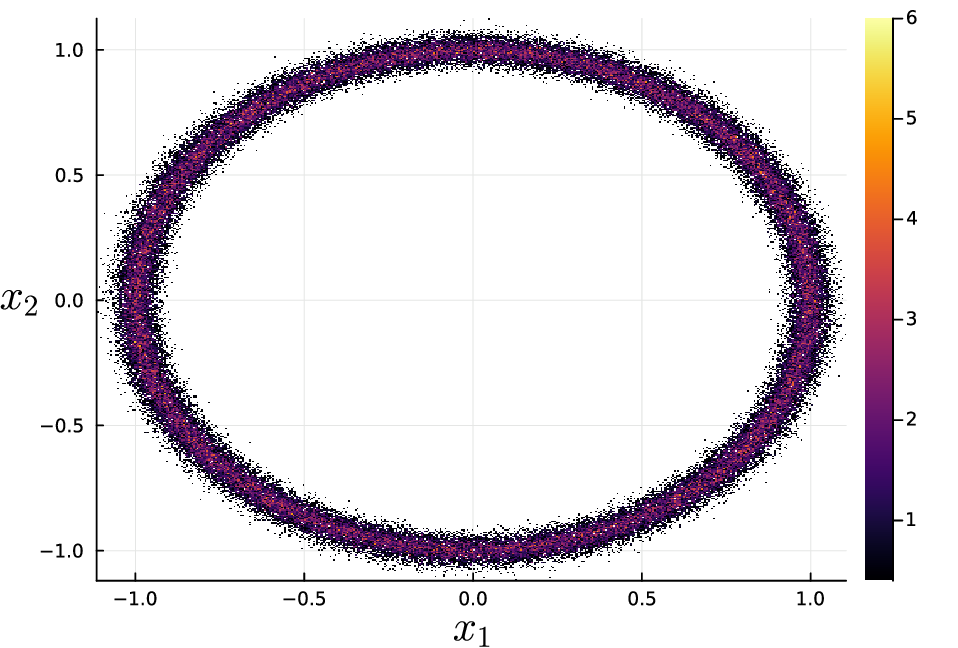}
    \subcaption{BPS}
    \label{fig:MGC_BPS_histogram}
  \end{minipage}
  \hspace{5mm}
  \begin{minipage}{.3\linewidth}
    \centering
    \includegraphics[width=\linewidth]{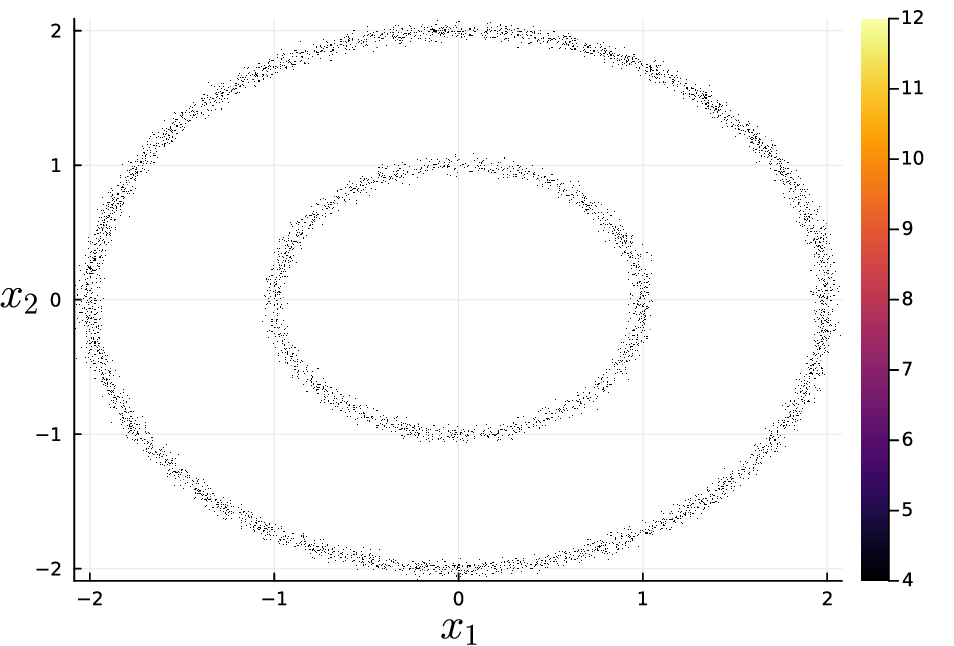}
    \subcaption{BPS-PT}
    \label{fig:MGC_BPS-PT_histogram}
  \end{minipage}
  \vspace{3mm}  
  \begin{minipage}{.3\linewidth}
    \centering
    \includegraphics[width=\linewidth]{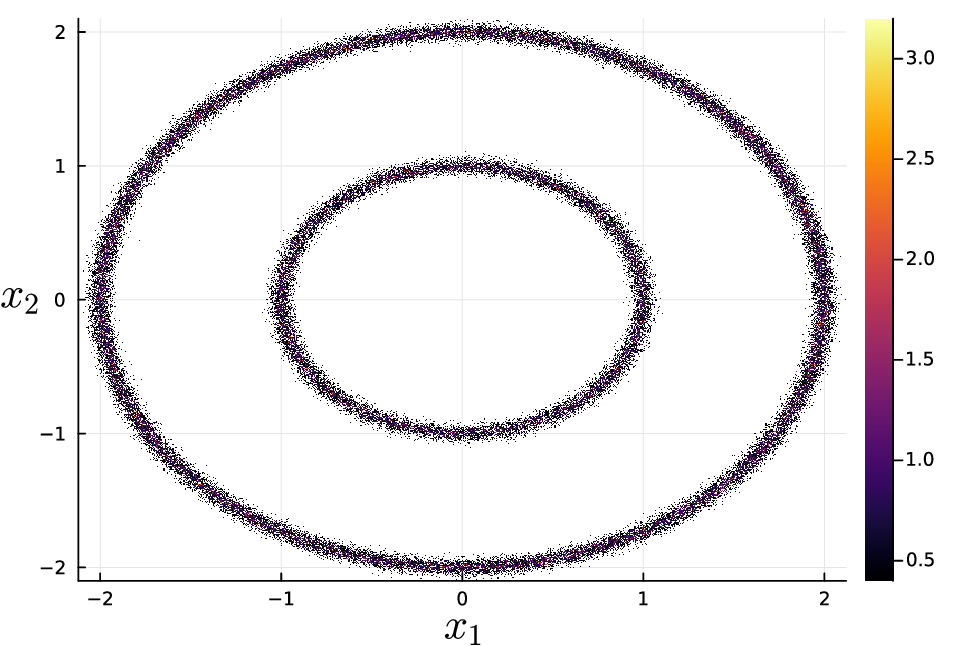}
    \subcaption{From the target pdf}
    \label{fig:MGC_pdf_histogram}
  \end{minipage}
 \caption{2D histograms of samples obtained by each method. To match the computation times, the number of samples was set to $10^4$ for BPS-PT, and $10^5$ for the others. Compared to the histogram obtained from the target pdf, each method had a following drawback. (a) exhibited anisotropy. (b) concentrated in the inner circumference than the outer one, which was opposite from the target pdf. (c) had no samples from the inner circumference. (d) had $1 / 10$ samples per computation time. 
 }
 \label{fig:MGC_histograms}
\end{figure}

Using an optimized parameter, we performed $10$ trials and obtained $10^4$ samples (BPS-PT) and $10^5$ samples (the other methods) for each trial (Tab.~\ref{tab:MGC_time}), respectively. 
The samples obtained from each method and the target pdf in Eq.~(\ref{eq:pdf_MGC}) are shown in Fig.~\ref{fig:MGC_histograms}. 
This indicates that HMC-PT, LMC-PT, and BPS-PT could obtain samples from both circumferences, however, there are significant drawbacks with the first two methods as noted in the figure caption. 
Performance metrics are shown in Tab.~\ref{tab:MGC_result}, where the value of KLD indicates that BPS-PT could estimate the ratio of the two distributions with the highest precision. 
Additionally, the ESS of BPS and BPS-PT was higher than that of HMC-PT and LMC-PT, which shows the advantage of BPS-based method. 
To summarize these results, we find that BPS-PT is the most useful method to obtain independent samples from MGC. 

The reason for advantage of BPS-based methods in ESS can be discussed as follows. 
For a distribution with large curvature change, $\Delta t_1$'s of HMC-PT and LMC-PT were taken to be small to keep optimal acceptance rates of MH step. 
However, to avoid a long computation time, the number of gradient updates should be limited, which made the autocorrelation slowly decrease. 
In BPS and BPS-PT, the particles moved along a circumference for a certain distance, and therefore the autocorrelation is considered to decay rapidly.

\begin{table}[ht]
 \centering
 \caption{Computation times of MGC are shown. The number of samples were $10^4$ for BPS-PT and $10^5$ for the other methods, respectively. To match the optimization time, the episode length for the policy gradient of BPS-PT is $1/10$ of that of HMC-PT and LMC-PT. The optimization time and sampling time are the average of $5$ and $10$ runs, respectively.}
 \label{tab:MGC_time}
\begin{tabular}{c|c|c|c|c}
     & HMC-PT & LMC-PT & BPS & BPS-PT \\ \hline
 Optimization [sec] & $6.3 \times 10^2$& $1.18 \times 10^3$ & - & $1.09 \times 10^3$ \\
 Sampling [sec]  & $1.44 \times 10^2$& $1.93 \times 10^2$ & $5 \times 10$ & $1 \times 10^2$ \\ 
\end{tabular}
\end{table}

\begin{table}[ht]
 \centering
 \caption{Performance metrics of MGC are shown. To match the computation times, BPS-PT used $10^4$ samples and the other method used $10^5$ samples, respectively. Since no transitions between two circumferences occurred by BPS, the KLD is infinity although sample means of $a$ was $0.7$.}
 \label{tab:MGC_result}
\begin{tabular}{c|c|c|c|c}
     & HMC-PT & LMC-PT & BPS & BPS-PT \\ \hline
 ESS & $2 \times 10^{-4}$& $0.2$ & $1.0$ & $1.0$ \\
 $W_1$  & $0.5$& $0.35$ & $0.8$ & $0.34$ \\ 
 KLD & $4 \times 10^{-2}$ & $1.3 \times 10^{-1}$ & $\infty$ & $2 \times 10^{-4}$ \\
 estimate of $a$ & $2.2 \times 10^{-2}$ & $5.5 \times 10^{-1}$ & $0.7$ & $3.0 \times 10^{-1}$ \\
\end{tabular}
\end{table}

 \subsection{Mixture of Neal's funnels}
The pdf of the Neal's funnel (NF)~\cite{neal2003slice} is given by 
\begin{align}
 p(x) = {\cal N}(x_1 \mid 0, {\rm e}^{x_2 / 2} ) \, {\cal N}(x_2 \mid 0, \sigma). 
\end{align}
Since the variance of $x_1$ changes exponentially as $x_2$ increases, 
$x_1$ is widely distributed for large $x_2$, while concentrated in a narrow region for small $x_2$.
Since the curvature of this model changes significantly, it is known that tuning a constant $\Delta t$ of HMC and LMC is difficult. 

To show the usefulness of the PT, we employed a mixture of NFs (MNF), 
\begin{align}
 p(x, y) =& \delta_{y, 0} \, a \, {\cal N}(x_1 \mid \mu, {\rm e}^{(x_2 - 2) / 2} ) \, {\cal N}(x_2 \mid 0, \sigma) \nonumber \\
 &+ \delta_{y, 1} \, (1 - a) \, {\cal N}(x_1 \mid - \mu, {\rm e}^{(- x_2 - 2) / 2} ) \, {\cal N}(x_2 \mid 0, \sigma), 
 \label{eq:pdf_NFM}
\end{align}
where we use $a = 0.3, \mu = -3, \sigma = 5$. 
The computation times to optimize the parameters and samplings are shown in Tab.~\ref{tab:NFMM_time}. 
Although the optimization time of HMC-PT seems to be shorter than that of the other methods, it should be noted that we excluded trials for which the calculation was not completed. 
This happens when a particle is in the tip part of a funnel and $\Delta t$ enormously decreased to optimize the acceptance ratio. 
A small $\Delta t$ also appeared in LMC-PT and resulted in a long optimization and sampling time. 
In contrast, BPS-PT does not include $\Delta t$, resulting in long but stable optimization times. 

Using an optimized parameter, we performed $10$ trials and obtained $10^6$ samples for each trial (Tab.~\ref{tab:NFMM_time}). 
Among $5$ optimization trials, we chose parameters of HMC-PT and LMC-PT so that sampling times became short. 
Although BPS did not calculate the temperature exchanges, it required more time for each update, resulting in a computation time similar to that of these two methods. 
Therefore, BPS-PT, which included the calculation of the temperature exchanges, took about $10$ times longer than the other methods. 

\begin{figure}[tbp]
  \centering
  \begin{minipage}{.3\linewidth}
    \centering
    \includegraphics[width=\linewidth]{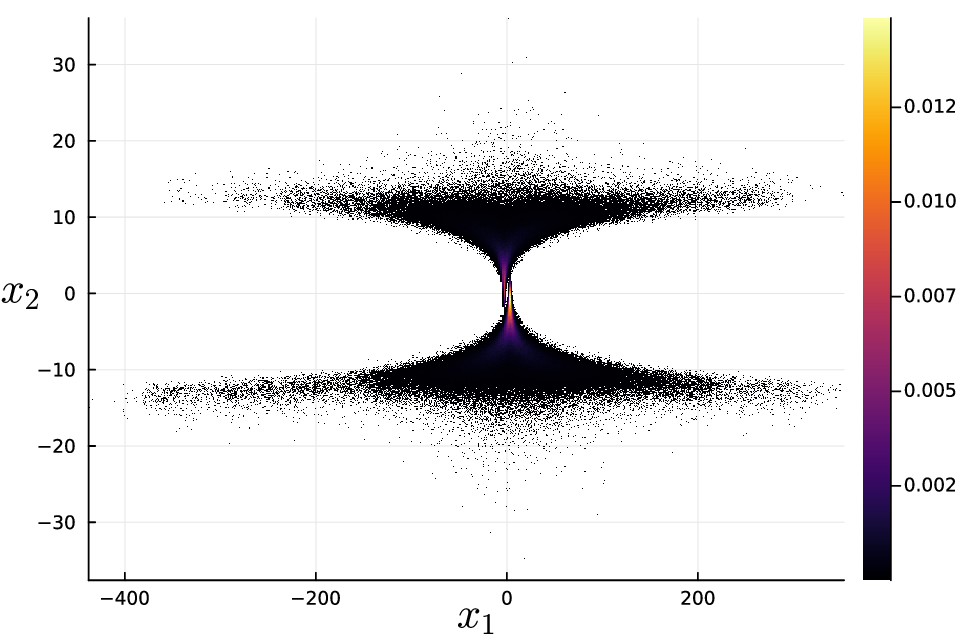}
    \subcaption{HMC-PT}
    \label{fig:neal_HMC_histogram}
  \end{minipage}
  \hspace{5mm}
  \begin{minipage}{.3\linewidth}
    \centering
    \includegraphics[width=\linewidth]{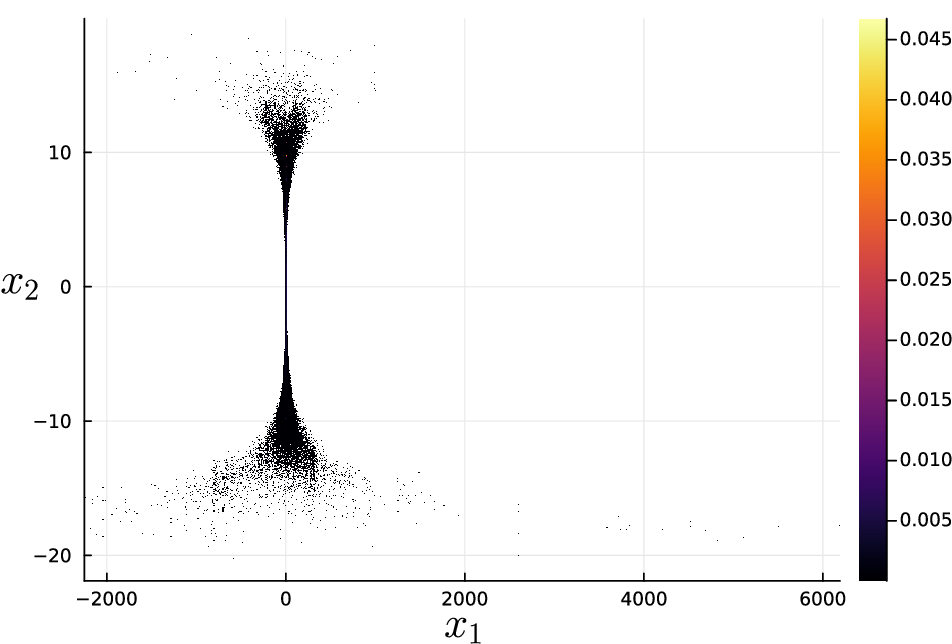}
    \subcaption{LMC-PT}
    \label{fig:neal_LMC_histogram}
  \end{minipage}
  \vspace{3mm}  
  \begin{minipage}{.3\linewidth}
    \centering
    \includegraphics[width=\linewidth]{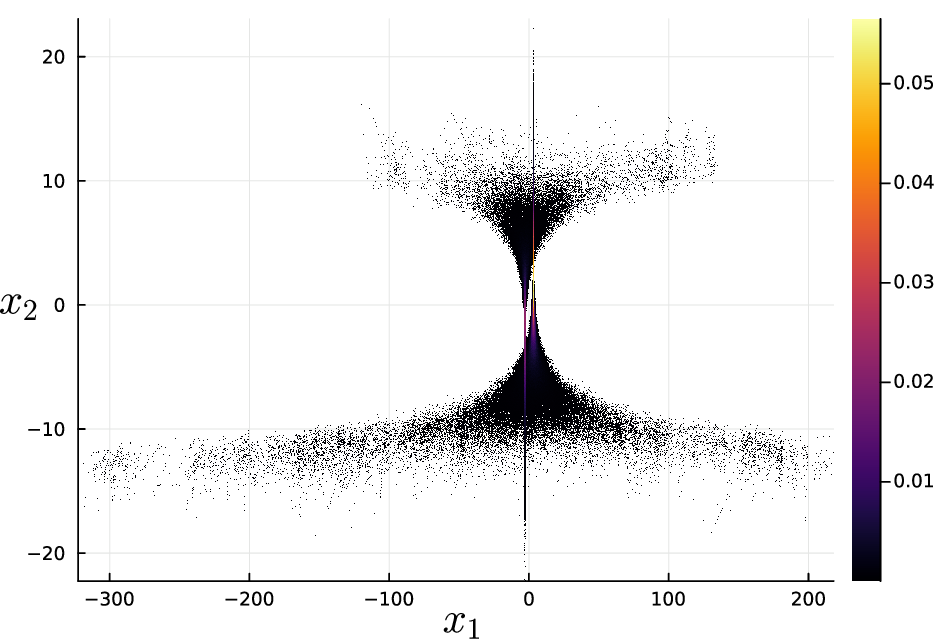}
    \subcaption{BPS}
    \label{fig:neal_BPS_histogram}
  \end{minipage}
  \hspace{5mm}
  \begin{minipage}{.3\linewidth}
    \centering
    \includegraphics[width=\linewidth]{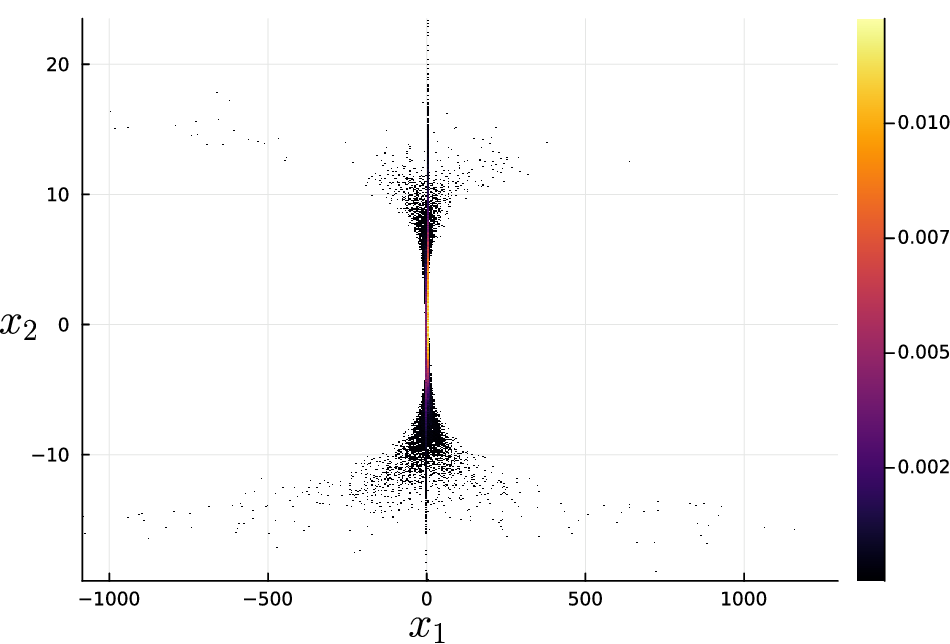}
    \subcaption{BPS-PT}
    \label{fig:neal_BPS-PT_histogram}
  \end{minipage}
  \vspace{3mm}  
  \begin{minipage}{.3\linewidth}
    \centering
    \includegraphics[width=\linewidth]{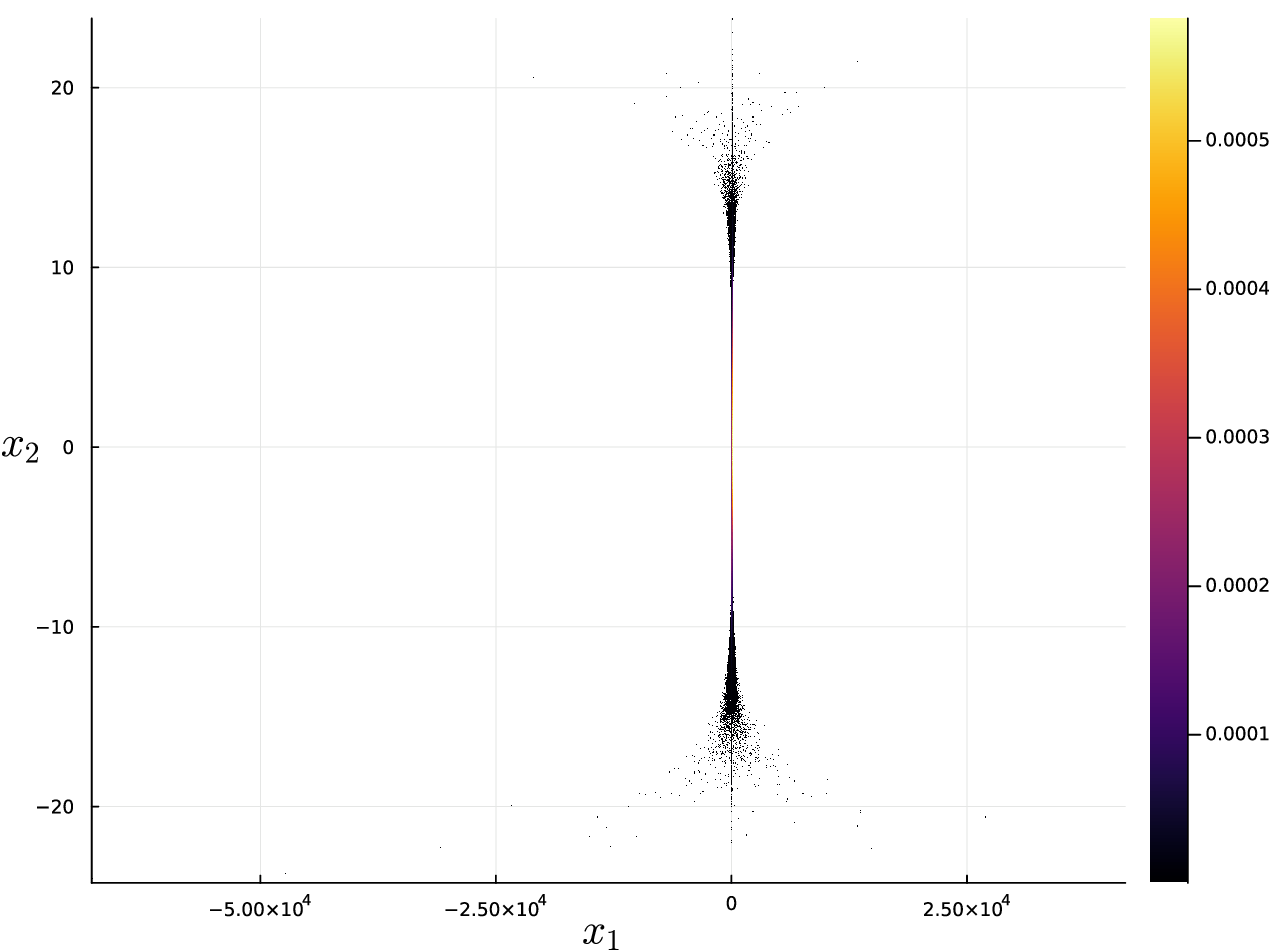}
    \subcaption{From the target pdf}
    \label{fig:neal_pdf_histogram}
  \end{minipage}
 \caption{2D histograms of samples obtained by each method. To match the computation times, the number of samples was set to $10^5$ for BPS-PT, and $10^6$ for the others, respectively. Compared to the histogram obtained from the target pdf, the horizontal spread ($x_1$) is narrower for all methods. The peaked part in the vertical direction ($x_2$) is reproduced more closely to the target pdf by BPS and BPS-PT than by HMC and LMC.
 }
 \label{fig:neal_histograms}
\end{figure}

The samples obtained from each method and the target pdf in Eq.~(\ref{eq:pdf_NFM}) are shown in Fig.~\ref{fig:neal_histograms}. 
(To match the computation time, $10^5$ samples were used for BPS-PT and $10^6$ samples for the other methods, respectively.) 
Compared to the histogram obtained from the target pdf, the horizontal spread ($x_2$) of the histograms for all methods is narrower since it is difficult for particles to move around in such a large area. 
From Fig.~\ref{fig:neal_pdf_histogram}, narrow vertical paths extended downward to $x_2 = -20$ for $y = 0$ and upward to $x_2 = 20$ for $y = 1$. 
This result is natural because we obtained $10^6$ samples from normal distribution and almost all samples will be observed in the range within $4 \sigma$, namely from $x_2 = -20$ to $x_2 = 20$ under $\sigma = 5$. 
The narrow vertical paths were reproduced more accurately by BPS and BPS-PT than by HMC-PT and LMC-PT, whose narrow paths ended before $\pm 20$. 
In BPS and BPS-PT, the particles could move vertically by repeatedly bouncing off the narrow potential walls. 
In HMC-PT and LMC-PT, to reduce sampling times, we used a discretized time $\Delta t$ optimized in regions where the curvature changes gently. 
However, such $\Delta t$'s were too rough to move the particles along the narrow paths in regions with high curvature changes.



Performance metrics are summarized in Tab.~\ref{tab:NFMM_result}. 
Although $W_1$ and KLD were slightly higher for HMC-PT, the values were roughly the same for all methods. 
Although the MNF is a multimodal distribution, transition from one mode to another frequently occurred. 
Hence, when sampling from the MNF, PT is considered to impose an excessive computational load, while PT is expected to improve ESS by exchanging particles located far apart. 
Nevertheless the ESSs of HMC-PT, LMC-PT and BPS were comparable, which indicates that BPS was useful to obtain independent samples. 
For BPS-PT, although the number of samples per computation time by BPS-PT was $1/10$ as those by the other methods, the ESS of BPS-PT was approximately $10^2$ times higher. 
Thus, we can consider that BPS-PT is superior to the other methods in terms of sampling independent particles from MNF. 

\begin{table}[ht]
 \centering
 \caption{Computation times of MNF are shown. The number of samples were $10^6$ for all methods. To match the optimization time, the episode length for the policy gradient of BPS-PT is $1/10$ of that of HMC-PT and LMC-PT. The optimization time and sampling time are the average of $5$ and $10$ runs, respectively.}
 \label{tab:NFMM_time}
\begin{tabular}{c|c|c|c|c}
     & HMC-PT & LMC-PT & BPS & BPS-PT \\ \hline
 Optimization [sec] & $8 \times 10$& $4 \times 10^2$ & - & $2.2 \times 10^2$ \\
 Sampling [sec]  & $6 \times 10$& $7.6 \times 10$ & $1.00 \times 10^2$ & $6.4 \times 10^2$ \\ 
\end{tabular}
\end{table}

\begin{table}[ht]
 \centering
 \caption{Performance metrics of MNF are shown. BPS-PT used $10^5$ samples and the other methods used $10^6$ samples to match computation times.}
 \label{tab:NFMM_result}
\begin{tabular}{c|c|c|c|c}
     & HMC-PT & LMC-PT & BPS & BPS-PT \\ \hline
 ESS & $1 \times 10^{-4}$& $2 \times 10^{-4}$ & $3 \times 10^{-4}$ & $4 \times 10^{-2}$ \\
 $W_1$ & $17$& $5$ & $4$ & $5$ \\ 
 KLD & $2.7 \times 10^{-3}$ & $2.2 \times 10^{-5}$ & $2.7 \times 10^{-6}$ & $1.5 \times 10^{-4}$ \\
 estimate of $a$ & $3.3 \times 10^{-1}$ & $3.0 \times 10^{-1}$ & $3.0 \times 10^{-1}$ & $2.9 \times 10^{-1}$ \\
\end{tabular}
\end{table}

 \subsection{Gaussian mixture model}
Next, we show an example of a target distribution with little curvature changes. 
The target distribution was the 24-dimensional 4-component Gaussian mixture model (GMM) used in Ref.~\cite{zhou2020mixed}, 
\begin{align}
 p(x, y) =& \sum_{i=1}^4 \phi_i \delta_{y,i} \, {\cal N}(x \mid \mu_i, \Sigma). 
 \label{eq:pdf_GMM}
\end{align}
The cluster probability was $\phi = (0.15, 0.3, 0.3, 0.25)$, the cluster centers were given by all permutations of $-2, 0, 2, 4$, and all covariance matrices were $\Sigma = 3I$. 
The computation times to parameter optimization and samplings are shown in Tab.~\ref{tab:GMM_time}. 
Since GMM in Eq.~(\ref{eq:pdf_GMM}) did not change the curvature, there is no need to change $\{\Delta t_i\}_{i=1}^L$ depending on the location. 
Thus, unlike MNF, GMM did not result in abnormally slow optimization times for HMC-PT and LMC-PT. 

Using an optimized parameter, we performed $10$ trials and obtained $10^6$ samples for each trial (Tab.~\ref{tab:GMM_time}). 
The computation time took about $10$ times longer for BPS-PT than the other methods. 
The performance metrics are summarized in Tab.~\ref{tab:GMM_result}. 
From $W_1$ and KLD, the methods other than BPS could correctly estimate the distribution. 
The large value of the KLD means BPS failed to estimate the cluster probability. 
It is shown that for a strongly multimodal distribution, transitions between the modes rarely occurred by BPS without PT. 
Since the cluster probability was not appropriately estimated, we did not evaluate ESS for BPS below. 
For a distribution with small curvature change, $\Delta t$'s were large compared to those with large curvature variations, which decreased the autocorrelation of samples. 
Additionally, by exchanging particles at high temperatures, the autocorrelation decreased more rapidly. 
As a result, high ESS values are obtained for HMC-PT and LMC-PT in Tab.~\ref{tab:GMM_result}. 
For this model, BPS-PT was inferior to HMC-PT and comparable to LMC-PT in terms of distribution estimation accuracy ($W_1$, KLD). 
Furthermore, the ESS of BPS-PT is lower than the other two, which means that BPS-PT is also inferior in terms of the number of independent samples per computation time in this model. 
Consequently, BPS and BPS-PT should not be used for distributions with small curvature changes like GMM. 


\begin{table}[ht]
 \centering
 \caption{Computation time of GMM are shown. The number of samples were $10^6$ for all methods. To match the optimization time, the episode length for the policy gradient of BPS-PT is $1/10$ of that of HMC-PT and LMC-PT. The optimization time and sampling time are the average of $5$ and $10$ runs, respectively.}
 \label{tab:GMM_time}
\begin{tabular}{c|c|c|c|c}
     & HMC-PT & LMC-PT & BPS & BPS-PT \\ \hline
 tuning [sec] & $1.09 \times 10^2$& $1.43 \times 10^2$ & - & $2.43 \times 10^2$ \\
 sampling [sec]  & $1.36 \times 10^2$& $1.21 \times 10^2$ & $8.1 \times 10$ & $1.21 \times 10^3$ \\ 
\end{tabular}
\end{table}

\begin{table}[ht]
 \centering
 \caption{Performance metrics of GMM are shown. To match calculation times, BPS-PT used $10^5$ samples and the other method used $10^6$ samples, respectively.}
 \label{tab:GMM_result}
\begin{tabular}{c|c|c|c|c}
     & HMC-PT & LMC-PT & BPS & BPS-PT \\ \hline
 ESS & $8 \times 10^{-4}$& $6.4 \times 10^{-2}$ & $5.8 \times 10^{-1}$ & $3 \times 10^{-4}$ \\
 $W_1$ & $3.35 \times 10$ & $3.59 \times 10$ & $6.4 \times 10$ & $3.30 \times 10$ \\ 
 KLD & $4.9 \times 10^{-4}$ & $2.7 \times 10^{-2}$ & $3.9$ & $2.3 \times 10^{-2}$ \\
\end{tabular}
\end{table}

\section{Conclusion}

MCMC~\cite{neal2011mcmc} is one of the most useful methods to obtain samples from complex distributions. 
For continuous distributions, HMC~\cite{neal2011mcmc} and LMC~\cite{rossky1978brownian} are widely used. 
These methods require to discretize time while keeping the detailed balance condition. 
For target distributions with large curvature change, a constant $\Delta t$ causes an inefficient sampling. 
In these cases, it is effective to adaptively tune $\Delta t$ by considering motion on a Riemannian manifold or transform distributions to ones with low curvature change. 
However, the former significantly increases computation time, and the latter may require to find nonlinear transformation. 
Recently, a sampling method called BPS was proposed, which uses uniform linear motion and elastic collisions with a potential~\cite{peters2012rejection, bouchard2018bouncy}. 
Since BPS does not discretize time, it is expected to be effective for target distributions with large curvature changes. 

In multimodal distributions, if transitions between modes rarely occur, convergence to the target distribution will be slow. 
PT was proposed to mitigate this problem~\cite{swendsen1986replica, neal1996sampling}. 
In LMC, it was shown that the higher the exchange rate of PT, the faster the convergence to the target distribution~\cite{dupuis2012infinite}. 
Then LMC-PT with the infinite exchange limit has been proposed~\cite{dupuis2012infinite}. 
Assuming that this also holds for BPS, we have proposed BPS-PT with infinite exchange rate in this article. 
In addition, we have extended the temperature optimization method using policy gradients for adjacent temperature exchange PT~\cite{zhao2024policy} to three or more temperature exchange PT. 

We applied HMC-PT, LMC-PT, BPS, and BPS-PT to distributions with large curvature changes (MGC and MNF) and small curvature changes (GMM). 
PT improved sampling efficiency for multimodal distributions in terms of both accuracy and independence. 
For multimodal distributions with large curvature change (MGC and MNF), BPS-PT was found to be superior to HMC-PT, LMC-PT, and BPS in terms of the number of independent samples obtained per computation time. 
In the case of a strongly multimodal distribution with small curvature change (GMM), BPS-PT was inferior even in terms of the number of independent samples per computation time, since the continuous-time simulation in BPS was slower than the gradient updates used in HMC-PT and LMC-PT. 
Thus, we can conclude that while BPS is effective for distributions with large curvature changes, it is best not to use it for distributions with small curvature changes. 
Unfortunately, for complex distributions, such as posterior distributions in Bayesian inference, the curvature change is unknown in advance. 
For this reason, it would be a good strategy to estimate the curvature changes using a fast sampling method such as HMC-PT or LMC-PT, and then use BPS-PT if necessary. 

\section*{Acknowledgments}
YS and KT were supported by KAKENHI No. 20H05776. 
KT was also supported by KAKENHI Nos. 20H05774, 23K10978, 25H01509, and a grant from Hagiwara Foundation of Japan.

\appendix
\section{Upper bounds of bounce rates}
 \subsection{Mixture of Gamma distributions with uniform circular distributions}
For simplicity, we deal with one component of the MGC. 
In the Cartesian coordinate, the pdf of a Gamma distribution with a circular uniform distribution is written by 
\begin{align}
 p(x) = \frac{1}{2 \pi \, \Gamma(k) \, \theta^k} \, \|x\|^{k - 2} {\rm e}^{- \|x\| / \theta}, 
\end{align}
where we omit the dependence of the parameter, $k, \theta$. 
Hence, the inner product of the bounce rate, $\lambda^{\rm b}$, is given by 
\begin{align}
 \left< - \nabla \ln p(x + v t), v\right>
 = \left( - \frac{k - 2}{\|x + v t\|^2} + \frac{1}{\theta \, \|x + v t\|} \right) \, \left< x + v t, v\right>.
 \label{eq:bounce_rate_circle}
\end{align}
To obtain the bounce rate, $\lambda^{\rm b}$, we determine the sign of Eq.~(\ref{eq:bounce_rate_circle}). 
First, the sign of the inner product changes at 
\begin{align}
 t_0 = - \frac{\left< x, v\right>}{\|v\|^2}. 
 \label{eq:circle_gamma_t_0}
\end{align}
Next, we define $D$ as 
\begin{align}
 D =& \frac{\left< x, v\right>^2}{\|v\|^4} - \frac{\|x\|^2 - \theta^2 (k - 2)^2}{ \|v\|^2}. 
 \label{eq:circle_gamma_D}
\end{align}
If $D \geq 0$, the sign of the terms in the bracket is negative in $t_0 - \sqrt{D} < t < t_0 + \sqrt{D}$, and if $D < 0$, the sign is positive. 
Then, we find that if $D \geq 0$, 
\begin{align}
 \lambda^{\rm b}(x + v t, v) 
 =& \max\left\{ 0, \left< - \nabla \ln p(x + v t), v\right> \right\} \nonumber \\
 =& \begin{cases}
  \left( - \frac{k - 2}{\|x + v t\|^2} + \frac{1}{\theta \, \|x + v t\|} \right) \, \left< x + v t, v\right> & (t_0 - \sqrt{D} \leq t \leq t_0, \, t \geq t_0 + \sqrt{D}) \\
  0 & ({\rm otherwise})
 \end{cases},
\end{align}
and if $D < 0$, 
\begin{align}
 \lambda^{\rm b}(x + v t, v) 
 =& \begin{cases}
  \left( - \frac{k - 2}{\|x + v t\|^2} + \frac{1}{\theta \, \|x + v t\|} \right) \, \left< x + v t, v\right> & (t \geq t_0) \\
  0 & (t < t_0)
 \end{cases}.
\end{align}
Applying the Cauchy-Schwarz's inequality to the inner product of $\lambda^{\rm b}$, 
\begin{align}
 - \|x + v t\| \, \|v\| \leq \left<x + v t, v\right> \leq \|x + v t\| \, \|v\|, 
\end{align}
we find that when $D \geq 0$, 
\begin{align}
 \lambda^{\rm b}(x + v t, v) 
 \leq& \begin{cases}
  \left( - \frac{k - 2}{\|x + v t\|} + \frac{1}{\theta} \right) \, \|v\| & (t \geq t_0 + \sqrt{D}) \\
  \left( \frac{k - 2}{\|x + v t\|} - \frac{1}{\theta} \right) \, \|v\| & (t_0 - \sqrt{D} \leq t \leq t_0) \\
  0 & ({\rm otherwise})
 \end{cases},
\end{align}
and if $D < 0$, 
\begin{align}
 \lambda^{\rm b}(x + v t, v) 
 \leq& \begin{cases}
  \left( - \frac{k - 2}{\|x + v t\|} + \frac{1}{\theta} \right) \, \|v\| & (t \geq t_0) \\
  0 & (t < t_0)
 \end{cases}.
\end{align}
The inflection points of the above upper bound are given by 
\begin{align}
 t_0 \pm \sqrt{\frac{\|x\|^2}{2 \|v\|^2} - \frac{t_0^2}{2}} =\colon t_0 \pm \alpha. 
\end{align}
Thus, using a numerical calculation, we can construct intervals with endpoints, $t_0, t_0 \pm \sqrt{D}, t_0 \pm \alpha$, and obtain an upper bound of $\lambda^{\rm b}$ in the form of a linear function.

 \subsection{Mixture of Neal's funnels}
Similarly to the MGC, for simplicity, we deal with one component of the MNF. 
The pdf of the NF is given by 
\begin{align}
 p(x) = {\cal N}(x_1 \mid \mu_1, {\rm e}^{(x_2 - \mu_2) / 2}) \, {\cal N}(x_2 \mid 0, \sigma_2).
\end{align}
The inner product is written by 
\begin{align}
 f_{x, v}(t) = \left< - \nabla \ln p(x + v t), v \right>
 =& {\rm e}^{- v_2 t} \, (a_0 + a_1 t + a_2 t^2) + b_0 + b_1 t, \\
 a_0 =& {\rm e}^{-x_2 + \mu_2} \, (x_1 - \mu_1) \, \left[ v_1 - \frac{v_2}{2} (x_1 - \mu_1) \right], \\
 a_1 =& {\rm e}^{-x_2 + \mu_2} \, v_1 \, [ v_1 - v_2 (x_1 - \mu_1) ], \\
 a_2 =& {\rm e}^{-x_2 + \mu_2} \, \left( - \frac{1}{2} v_1^2 v_2 \right), \\
 b_0 =& \left( \frac{1}{2} + \frac{x_2}{\sigma_2^2} \right) v_2, \\
 b_1 =& \frac{v_2^2}{\sigma_2^2}.
\end{align}
Since the second derivative of $f_{x, v}$ can be expressed as 
\begin{align}
 f''_{x, v}(t) =& {\rm e}^{- v_2 t} \, g_{x, v}(t), \\
 g_{x, v}(t) =& (v_2^2 a_0 - 2 v_2 a_1 + 2 a_2) + (v_2^2 a_1 - 4 v_2 a_2) t + v_2^2 a_2 t^2, 
\end{align}
we can find the inflection points by solving $g_{x, v}(t) = 0$. 
Thus, we obtain an upper bound of the bounce rate, $\lambda^{\rm b}$, in the form of a linear function.


\end{document}